\documentclass{amsart}

\theoremstyle{definition}

\theoremstyle{remark}

\numberwithin{equation}{section}

\begin{document}

\title[A three layer neural network can represent any function]{A three layer neural network can represent any
multivariate function}

\author{Vugar E. Ismailov}

\address{Institute of Mathematics and Mechanics, Azerbaijan National Academy of Sciences, 9 B.~Vahabzadeh str., AZ1141, Baku, Azerbaijan}

\email{vugaris@mail.ru}



\begin{abstract}
In 1987, Hecht-Nielsen showed that any continuous
multivariate function can be implemented by a certain type three-layer
neural network. This result was very much discussed in neural network
literature. In this paper we prove that not only continuous functions but
also all discontinuous functions can be implemented by such neural networks.
\end{abstract}

\maketitle

\section{Introduction}
Over the past 30 years, the topic of artificial neural networks has been a vibrant
area of research. Neural networks are powerful computation devices, which have applications in many fields
and problem domains. Application areas range from medicine to petroleum science and geology.
In fact neural networks are introduced in any situation where there are problems of prediction, classification or control.
Undoubtedly the greatest advantage of neural networks is their ability to be
used as an arbitrary function approximation and/or implementation mechanism. In the present article, we are
interested in the question of precise representation of multivariate functions by neural networks.

It should be remarked that one of the pioneering papers in neural network theory is the 1987 paper by Hecht-Nielsen \cite{Hecht}.
This paper gained tremendous attention of many researchers during
the following decades after its publication. The main result of \cite{Hecht}
is based on the Kolmogorov Superposition Theorem \cite{Kol}. It shows that
any continuous multivariate function can be implemented by a special three-layer
neural network as follows.

\bigskip

\textbf{Theorem 1} (Hecht-Nielsen \cite{Hecht}). \textit{Given a natural number $d>1$ and
any continuous function $f:\mathbb{I}^{d}\rightarrow \mathbb{R}$, $y=f(%
\mathbf{x})$, where $\mathbb{I}$ is the closed unit interval $[0,1].$ Then $%
f $ can be implemented exactly by three-layer neural network having $d$
processing elements $\mathbf{x}=(x_{1},...,x_{d})$ in the first (input)
layer, $2d+1$ processing elements in the middle layer and a processing
element $y$ in the top (output) layer.}

\textit{The processing elements on the first layer simply distribute the
input $\mathbf{x}$-vector components to the processing elements of the
second layer.}

\textit{The processing elements of the second layer implement the following
transfer functions:}
\begin{equation*}
z_{k}=\sum_{j=1}^{d}\lambda ^{j-1}\phi (x_{j}+\epsilon k),\text{ }k=0,...,2d,
\end{equation*}%
\textit{where $\phi $ is a universal monotonic Lipschitz function and $%
\lambda ,\epsilon $ are nonzero constants. These $\phi ,\lambda ,$ and $%
\epsilon $ are independent of $f$. Moreover, the constant $\epsilon $ can be
chosen arbitrarily.}

\textit{The top layer processing element $y$ have the following transfer
function:}

\begin{equation*}
y=\sum_{k=0}^{2d}g_{k}(z_{k}),
\end{equation*}%
\textit{where the functions $g_{k}$ are real continuous and depend on $f$.}

\bigskip

This result is a neural network interpretation of Kolmogorov's superposition
theorem in the form given by Sprecher in \cite{Sp72}. For a comprehensive
discussion of this remarkable theorem see the book by Khavinson \cite{Kh}. A neural network, existence
of which is asserted in Theorem 1, is called Kolmogorov's mapping neural network.

Note that the original theorem of Hecht-Nielsen uses Sprecher's earlier
result from \cite{Sp65}. We have formulated this theorem using Sprecher's
later result \cite{Sp72}, which was also extensively discussed in a number
of subsequent papers (see, e.g., \cite{Br,Kop,Sp93,Sp96,Sp97}).

Although the above theorem completely characterizes the power of feedforward
neural networks, it was considered by some authors as non-constructive (see,
e.g., \cite{Gir}). Hecht-Nielsen himself wrote in \cite{Hecht} that the
theorem is strictly an existence result, this theorem tells us that such a
three-layer mapping network must exist, but it doesn't tell us how we can
construct it. Nevertheless, Hecht-Nielsen's theorem stimulated the further
research concerning the role of Kolmogorov superposition theorem in neural
network theory, which is still active today (see, e.g., \cite{Jorg,Mont,Sch,Shen}).
The research on this subject was carried out mainly in two
directions. In the first direction, the analysis was concentrated on
approximative versions of Kolmogorov's theorem and similar results on
feedforward neural networks (see, e.g., \cite{GI,Is14,Notes,Kur1,Kur2,MP,Pin}%
). In the second direction, the precise form of the Kolmogorov superposition
theorem and its relationship to neural networks were studied.
Hecht-Nielsen's expectations that \textquotedblleft more will be learned
about the Kolmogorov mapping network in the years to come" were met by
series of works of Sprecher \cite{KS94,Sp93,Sp96,Sp97}. Due to these works,
there is a perspective for a practical usage of the exact representation of
continuous functions by Kolmogorov type neural networks. In fact, such a
perspective stems from the fact that Sprecher's function $\phi $, which
determines the processing units of the middle layer, could be computed
algorithmically (see, e.g., \cite{Br,Kop,Sp96}).

Note that the external activation functions $g_{k}$ in the output layer depend on $f$
and have to be determined by learning procedures. Some practically useful
learning algorithms for such networks were discussed in \cite{Ner}. By using
cubic spline technique of approximation, both for external and internal functions in Kolmogorov type networks, more efficient approximation
of multivariate functions was achieved (see \cite{Igel}). In \cite{Brat},
Brattka obtained a computable version of Hecht-Nielsen's theorem:
Every computable multivariate function can be implemented by a three-layer neural network
with computable activation functions and computable
weights.

It is well known that in nature most functional dependencies are not
continuous. Regarding complicated discontinuous functional dependencies the
following question arises: can discontinuous functions be implemented by
Kolmogorov type neural networks? In this paper we prove that the answer to
this fair and interesting question is positive. More precisely, we prove
that Hecht-Nielsen's theorem can be extended from the class of continuous
multivariate functions to the class of all multivariate functions.

\section{Main result}
The following theorem is valid.

\bigskip

\textbf{Theorem 2.} \textit{Given a natural number $d>1$ and any function $F:%
\mathbb{I}^{d}\rightarrow \mathbb{R}$, $y=F(\mathbf{x})$, where $\mathbb{I}$
is the closed unit interval $[0,1].$ Then $F$ can be implemented exactly by
three-layer neural network having $d$ processing elements $\mathbf{x}%
=(x_{1},...,x_{d})$ in the first layer, $2d+1$ processing elements in the
middle layer and a single processing element $y$ in the top layer.}

\textit{The processing elements on the first layer simply distribute the
input $\mathbf{x}$-vector components to the processing elements of the
second layer.}

\textit{The processing elements of the second layer implement the following
transfer functions:}
\begin{equation*}
z_{k}=\sum_{j=1}^{d}\lambda ^{j-1}\phi (x_{j}+\epsilon k),\text{ }k=0,...,2d,
\end{equation*}%
\textit{where the universal monotonic Lipschitz function $\phi $ and the
nonzero constants $\lambda ,\epsilon $ are independent of $F$. Moreover, the
constant $\epsilon $ can be chosen arbitrarily.}

\textit{The top layer processing element $y$ have the following transfer
function:}

\begin{equation*}
y=\sum_{k=0}^{2d}h_{k}(z_{k}),
\end{equation*}%
\textit{where the functions $h_{k}$ are real and depend on $F$.}

\bigskip

One can observe that the only difference between conclusions of Theorems 1
and 2 are in functions $g_{k}$ and $h_{k}$. Sprecher's function $\phi $ and
the constants $\lambda ,\epsilon $ and the number of layers and units in
these layers are the same in both theorems. Based on Theorem 2, we can say
that not only continuous functions but also all discontinuous functions can
be implemented by Kolmogorov's mapping neural network. Our proof is based on
methods and principles of Functional Analysis.

\bigskip

\textbf{Proof.} For simplicity of notation put $r=2d$. Note that by Theorem
1 every continuous function $f:\mathbb{I}^{d}\rightarrow \mathbb{R}$ has the
form
\begin{equation*}
f(\mathbf{x})=\sum_{k=0}^{r}g_{k}(z_{k}(\mathbf{x})).\eqno(1)
\end{equation*}%
Using this, we show below that the family of the transfer functions $%
z_{0},...,z_{r}$ satisfies the following condition, which we call Condition
(Z):

\bigskip

\textit{Condition (Z):} There is no finite subset $\{\mathbf{x}_{1},...,%
\mathbf{x}_{n}\}\subset $ $\mathbb{I}^{d}$ with the property that

\begin{equation*}
\sum_{j=1}^{n}\mu _{j}\delta _{z_{k}(\mathbf{x}_{j})}(t)=0,~k=0,...,r,\eqno%
(2)
\end{equation*}%
for some nonzero real numbers $\mu _{1},...,\mu _{n}$ and for any $t\in
\mathbb{R}$. Here $\delta _{a}$ stands for the indicator function of a
single point set $\{a\}$. That is,

\begin{equation*}
\delta _{a}(t)=\left\{
\begin{array}{c}
1,\text{ if }t=a \\
0,\text{ if }t\neq a%
\end{array}%
\right. .
\end{equation*}

\bigskip

Let us first explain Eq. (2) in detail. We see that it stands for a system
of ceratin linear equations. Fix the subscript $k.$ Let the set $\{z_{k}(%
\mathbf{x}_{j}),$ $j=1,...,n\}$ have $s_{k}$ different values, which we
denote by $\gamma _{1}^{k},\gamma _{2}^{k},...,\gamma _{s_{k}}^{k}.$ Take
the first number $\gamma _{1}^{k}.$ Putting $t=\gamma _{1}^{k}$, we obtain
from (2) that

\begin{equation*}
\sum_{j}\mu _{j}=0,
\end{equation*}%
where the sum is taken over all $j$ such that $z_{k}(\mathbf{x}_{j})=\gamma
_{1}^{k}.$ This is the first linear equation in $\mu _{j}$ corresponding to $%
\gamma _{1}^{k}$. Take now $\gamma _{2}^{k}$. By the same way, putting $%
t=\gamma _{2}^{k}$ in (2), we can form the second equation. Continuing until
$\gamma _{s_{k}}^{k}$, we obtain $s_{k}$ linear homogeneous equations in $%
\mu _{1},...,\mu _{n}$. The coefficients of these equations are the integers
$0$ and $1$. By varying $k$, we finally obtain $s=\sum_{k=0}^{r}s_{k}$ such
equations. Hence (2), in its expanded form, stands for the system of these
linear equations. Thus Condition (Z) means that no system of linear
equations of the form (2) has a solution with nonzero components.

It should be remarked that finite sets $\{\mathbf{x}_{1},...,\mathbf{x}%
_{n}\} $ satisfying (2) with respect to not only the transfer functions $%
z_{k}$ but arbitrary multivariate functions were exploited under the name of
\textquotedblleft closed paths" in several papers of the author (see, e.g.,
\cite{Is08,Is12,Is17}).

Let us now show that if representation (1) is valid for every continuous $f$%
, then Condition (Z) holds. Assume the contrary. Assume that there is a
finite set $p=\{\mathbf{x}_{1},...,\mathbf{x}_{n}\}$ in $\mathbb{I}^{d}$
with the property (2). Consider the following linear functional

\begin{equation*}
G_{p}(f)=\sum_{j=1}^{n}\mu _{j}f(\mathbf{x}_{j}).
\end{equation*}

It is not difficult to see that this functional annihilates all sums of the form $\sum_{k=0}^{r}g_k(z_{k}(\mathbf{x}))$, and hence, by the representation (1),
every continuous function $f$ on $\mathbb{I}^{d}$. That is, $G_{p}(f)=0$ for
any $f\in C(\mathbb{I}^{d})$. On the other hand by Urysohn's well-known
lemma (see, e.g., \cite{KF}) there exists a continuous function $f_{0}$ with
the property: $f_{0}(\mathbf{x}_{j})=1$ for indices $j$ such that $\mu
_{j}>0 $; $f_{0}(\mathbf{x}_{j})=-1$ for indices $j$ such that $\mu _{j}<0$;
and $-1<f_{0}(\mathbf{x})<1 $ for $\mathbf{x}\in \mathbb{I}^{d}\setminus p$.
For this function $G_{p}(f_{0})=\sum_{j=1}^{n}\left\vert \mu _{j}\right\vert
\neq 0.$ The obtained contradiction means that Condition (Z) holds for the
transfer functions $z_{0},...,z_{r}$.

Now we are going to prove that if Condition (Z) holds for any family of real
functions $w_{k}:\mathbb{I}^{d}\rightarrow \mathbb{R}$, $k=0,...,r$, with
pairwise disjoint ranges, then any multivariate (not necessarily continuous)
function $F:\mathbb{I}^{d}\rightarrow \mathbb{R}$, $y=F(\mathbf{x})$,
possess the representation
\begin{equation*}
F(\mathbf{x})=\sum_{k=0}^{r}s_{k}(w_{k}(\mathbf{x})),\eqno(3)
\end{equation*}%
where the functions $s_{k}:\mathbb{R}\rightarrow \mathbb{R}$ depend on $F$.

Introduce the notation
\begin{eqnarray*}
Y_{k} &=&w_{k}(\mathbb{I}^{d}),~k=0,...,r; \\
\Omega &=&Y_{0}\cup ...\cup Y_{r}.
\end{eqnarray*}%
By our assumption, $Y_{k}$, the ranges of the $w_{k}$, are pairwise disjoint
sets. That is, $Y_{i}\cap Y_{j}=\emptyset$, for all $i,j\in \{0,...,r\},$
$i\neq j$.

Consider the following set
\begin{equation*}
\mathcal{L}=\{Y=\{y_{0},...,y_{r}\}:\text{if there exists }\mathbf{x}\in
\mathbb{I}^{d}\text{ s.t. }w_{k}(\mathbf{x})=y_{k},~k=0,...,r\}\eqno(4)
\end{equation*}%
Note that $\mathcal{L}$ is not a subset of $\Omega $. It is a set of some
certain subsets of $\Omega .$ Each element of $\mathcal{L}$ is a set $%
Y=\{y_{0},...,y_{r}\}\subset \Omega $ with the property that there exists at
least one point $\mathbf{x}\in \mathbb{I}^{d}$ such that $w_{k}(\mathbf{x}%
)=y_{k},~k=0,...,r.$ These $\mathbf{x}$ will be called generating points for
$Y$.

It is not difficult to understand that in (4) for each element $Y$ there
exists only one point $\mathbf{x}\in \mathbb{I}^{d}$. This is because if
there are two points $\mathbf{x}_{1}$ and $\mathbf{x}_{2}$ for a single $Y$
in (4), then $w_{k}(\mathbf{x}_{1})=w_{k}(\mathbf{x}_{2})$, $k=0,...,r$, and
hence for the set $\left\{ \mathbf{x}_{1},\mathbf{x}_{2}\right\} \subset
\mathbb{I}^{d}$ we have
\begin{equation*}
1\cdot \delta _{w_{k}(\mathbf{x}_{1})}+(-1)\cdot \delta _{w_{k}(\mathbf{x}%
_{2})}\equiv 0,~k=0,...,r.
\end{equation*}%
But this contradicts the assumption that Condition (Z) holds for the
functions $w_{0},...,w_{r}$. Thus we see that the unicity property of
generating points holds for each $Y$ in $\mathcal{L}$.

Since we already know that in (4) for each $Y\in \mathcal{L}$ there exists
only one point $\mathbf{x}\in \mathbb{I}^{d}$, we can define the function
\begin{equation*}
t:\mathcal{L}\rightarrow \mathbb{R},~t(Y)=F(\mathbf{x}),
\end{equation*}%
where $\mathbf{x}$ is the generating point for $Y$.

Consider now a class $\mathcal{S}$ of functions of the form $%
\sum_{j=1}^{m}r_{j}\delta _{D_{j}},$ where $m$ is a positive integer, $r_{j}$
are real numbers and $D_{j}$ are elements of $\mathcal{L},~j=1,...,m.$ We
fix neither the numbers $\ m,~r_{j},$ nor the sets $D_{j}.$ Clearly, $%
\mathcal{S\ }$is a linear space. Over $\mathcal{S}$, we define the functional
\begin{equation*}
H:\mathcal{S}\rightarrow \mathbb{R},~H\left( \sum_{j=1}^{m}r_{j}\delta
_{D_{j}}\right) =\sum_{j=1}^{m}r_{j}t(D_{j}).
\end{equation*}

First of all, we must show that this functional is well defined. That is,
once we have the equality%
\begin{equation*}
\sum_{j=1}^{m_{1}}r_{j}^{\prime }\delta _{D_{j}^{\prime
}}=\sum_{j=1}^{m_{2}}r_{j}^{\prime \prime }\delta _{D_{j}^{\prime \prime }},%
\eqno(5)
\end{equation*}%
we also have the equality

\begin{equation*}
\sum_{j=1}^{m_{1}}r_{j}^{\prime }t(D_{j}^{\prime
})=\sum_{j=1}^{m_{2}}r_{j}^{\prime \prime }t(D_{j}^{\prime \prime }).
\end{equation*}%
But in fact equality (5) can never hold. Suppose the contrary. Suppose that
(5) holds. Let us write (5) in the equivalent form

\begin{equation*}
\sum_{j=1}^{m}r_{j}\delta _{D_{j}}=0.\eqno(6)
\end{equation*}%
Each set $D_{j}$ consists of $r+1$ real numbers $y_{0}^{j},...,y_{r}^{j}$, $%
j=1,...,m.$ By our assumption concerning the ranges of the $w_{k}(\mathbf{x}%
) $, all these numbers are different. Therefore,

\begin{equation*}
\delta _{D_{j}}=\sum_{i=0}^{r}\delta _{y_{i}^{j}},~j=1,...,m.\eqno(7)
\end{equation*}%
Eq. (7) together with Eq. (6) give

\begin{equation*}
\sum_{i=0}^{r}\sum_{j=1}^{m}r_{j}\delta _{y_{i}^{j}}=0.\eqno(8)
\end{equation*}%
Since the sets $\{y_{i}^{1},y_{i}^{2},...,y_{i}^{m}\}$, $i=0,...,r,$ are
pairwise disjoint, we obtain from (8) that
\begin{equation*}
\sum_{j=1}^{m}r_{j}\delta _{y_{i}^{j}}=0,\text{ }i=0,...,r.\eqno(9)
\end{equation*}

Let now $\mathbf{x}_{1},...,\mathbf{x}_{m}$ be generating points for the
sets $D_{1},...,D_{m}$, respectively. Since by (4), $y_{k}^{j}=w_{k}(\mathbf{%
x}_{j})$, for $k=0,...,r,$ and $j=1,...,m,$ it follows from (9) that the set
$\{\mathbf{x}_{1},...,\mathbf{x}_{m}\}$ has property (2), hence Condition
(Z) is violated. The obtained contradiction means that Eq. (6), hence Eq.
(5) can never hold. Thus, the functional $H$ is well defined. Note that this
functional is linear, which can easily be seen from its definition.

Consider now the following space:

\begin{equation*}
\mathcal{S}^{\prime }=\left\{ \sum_{j=1}^{m}r_{j}\delta _{\omega
_{j}}\right\} ,
\end{equation*}%
where $m\in \mathbb{N}$, $r_{j}\in \mathbb{R}$, $\omega _{j}\subset \Omega .$
As above, we do not fix the parameters $m$, $r_{j}$ and $\omega _{j}.$
Clearly, the space $\mathcal{S}^{\prime }$ is larger than $\mathcal{S}$. Let
us prove that the functional $H$ can be linearly extended to the space $%
\mathcal{S}^{\prime }$. So, we must prove that there exists a linear
functional $H^{\prime }:\mathcal{S}^{\prime }\rightarrow \mathbb{R}$ such
that $H^{\prime }(x)=H(x)$, for all $x\in \mathcal{S}$. Let $A$ denote the
set of all linear extensions of $H$ to subspaces of $\mathcal{S}^{\prime }$
containing $\mathcal{S}$. The set $A$ is not empty, since it contains the
functional $H.$ For each functional $v\in A$, let $dom(v)$ denote the domain
of $v$. Consider the following partial order in $A$: $v_{1}\leq v_{2}$, if $%
v_{2}$ is a linear extension of $v_{1}$ from the space $dom(v_{1})$ to the
space $dom(v_{2}).$ Let now $P$ be any chain (linearly ordered subset) in $A$%
. Consider the following functional $u$ defined on the union of domains of
all functionals $p\in P$:
\begin{equation*}
u:\bigcup\limits_{p\in P}dom(p)\rightarrow \mathbb{R},~u(x)=p(x),\text{ if }%
x\in dom(p).
\end{equation*}

Obviously, this functional is well defined and linear. Besides, the
functional $u$ provides an upper bound for $P.$ We see that the arbitrarily
chosen chain $P$ has an upper bound. Then by Zorn's lemma (see, e.g., \cite%
{KF}), there is a maximal element $H^{\prime }\in A$. We claim that the
functional $H^{\prime }$ must be defined on the whole space $\mathcal{S}%
^{\prime }$. Indeed, if $H^{\prime }$ is defined on a proper subspace $%
\mathcal{D\subset }$ $\mathcal{S}^{\prime }$, then it can be linearly
extended to a space larger than $\mathcal{D}$ by the following way: take any
point $x\in \mathcal{S}^{\prime }\backslash \mathcal{D}$ and consider the
linear space $\mathcal{D}^{\prime }=\{\mathcal{D}+\alpha x\}$, where $\alpha
$ runs through all real numbers. For an arbitrary point $y+\alpha x\in
\mathcal{D}^{\prime }$, set $H^{^{\prime \prime }}(y+\alpha x)=H^{\prime
}(y)+\alpha b$, where $b$ is any real number considered as the value of $%
H^{^{\prime \prime }}$ at $x$. Thus, we constructed a linear functional $%
H^{^{\prime \prime }}\in A$ satisfying $H^{\prime }\leq H^{^{\prime \prime
}}.$ The last contradicts the maximality of $H^{\prime }.$ This means that
the functional $H^{\prime }$ is defined on the whole $\mathcal{S}^{\prime }$
and $H\leq H^{\prime }$ ($H^{\prime }$ is a linear extension of $H$).

Define the following functions by means of the functional $H^{\prime }$:
\begin{equation*}
s_{k}:Y_{k}\rightarrow \mathbb{R},\text{ }s_{k}(y_{k})\overset{def}{=}%
H^{\prime }(\delta _{y_{k}}),\text{ }k=0,...,r.
\end{equation*}%
Let $\mathbf{x}$ be an arbitrary point in $\mathbb{I}^{d}.$ Obviously, $%
\mathbf{x}$ is a generating point for some set $Y=\{y_{0},...,y_{r}\}\subset
\mathcal{L}.$ Thus,
\begin{equation*}
F(\mathbf{x})=t(Y)=H(\delta _{Y})=H\left( \sum_{k=0}^{r}\delta
_{y_{k}}\right) =H^{\prime }\left( \sum_{k=0}^{r}\delta _{y_{k}}\right) =
\end{equation*}

\begin{equation*}
\sum_{k=0}^{r}H^{\prime }(\delta
_{y_{k}})=\sum_{k=0}^{r}s_{k}(y_{k})=\sum_{k=0}^{r}s_{k}(w_{k}(\mathbf{x})).
\end{equation*}

Thus we have proven (3). We stress again that Eq. (3) is valid for any
family of real functions $w_{k}:\mathbb{I}^{d}\rightarrow \mathbb{R}$, $%
k=0,...,r$, which have disjoint ranges and satisfy Condition (Z).

Let us now return to the transfer functions $z_{k}$. Consider a system of
intervals $\{(a_{k},b_{k})\subset \mathbb{R}\}_{k=0}^{r}$ such that $%
(a_{i},b_{i})\cap (a_{j},b_{j})=\emptyset$ for all the indices $i,j\in
\{0,...,r\}$, $~i\neq j.$ For $k=0,...,r$, let $\tau _{k}$ be one-to-one
mappings of $\mathbb{R}$\ onto $(a_{k},b_{k}).$ Introduce the following
functions on $\mathbb{I}^{d}$:
\begin{equation*}
w_{k}(\mathbf{x})=\tau _{k}(z_{k}(\mathbf{x})),\text{ }k=0,...,r.
\end{equation*}

It is clear that (2) holds if and only if

\begin{equation*}
\sum_{j=1}^{n}\mu _{j}\delta _{w_{k}(\mathbf{x}_{j})}=0,~k=0,...,r.
\end{equation*}%
This means that in addition to the transfer functions $z_{0},...,z_{r}$,
Condition (Z) are also valid for the functions $w_{0},...,w_{r}$.

Note that these new functions $w_{k}(\mathbf{x})$ have pairwise disjoint
ranges. That is, $w_{i}(\mathbb{I}^{d})\cap w_{j}(\mathbb{I}%
^{d})=\emptyset,$ for all $i,j\in \{0,...,r\},~i\neq j.$ Then by Eq. (3)
we can write that
\begin{equation*}
F(\mathbf{x})=\sum_{k=0}^{r}s_{k}(w_{k}(\mathbf{x}))=\sum_{k=0}^{r}s_{k}(%
\tau _{k}(z_{k}(\mathbf{x})))=\sum_{k=0}^{r}h_{k}(z_{k}(\mathbf{x})),\eqno%
(10)
\end{equation*}%
where $h_{k}=s_{k}\circ \tau _{k}$, $k=0,...r$, are real univariate
functions depending on $F$. The obtained Eq. (10) proves the theorem.

\section{Conclusion}
Most multivariate functions that exist in nature and we see in practice are
generally not continuous. Although artificial neural networks were proved by
many authors to have the capability of representing and approximating all
continuous functions, their power to characterize discontinuous functions
was not known. This paper shows that Kolmogorov's mapping three-layer neural
network can precisely represent all discontinuous multivariate functions.

It should be remarked that Theorem 2, like Hecht-Nielsen's theorem, is
strictly an existence result. It only states that neural networks
implementing discontinuous functions exist and can be obtained by using
Kolmogorov's superposition theorem for continuous functions. The direct
application of Theorem 2 to practical problems is doubtful, since our method
for determining the functions $h_{k}$ is substantially based on Zorn's lemma
and hence highly nonconstructive. However, we hope that efficient learning
algorithms for such networks will be developed in the future.

\

\end{document}